\documentclass[letterpaper]{article} 
\usepackage{aaai2026}  
\usepackage{times}  
\usepackage{helvet}  
\usepackage{courier}  
\usepackage[hyphens]{url}  
\usepackage{graphicx} 
\usepackage{natbib}  
\usepackage{caption} 
\usepackage{algorithm}
\usepackage{algorithmic}
\usepackage{amsmath}

\usepackage{newfloat}
\usepackage{listings}
\DeclareCaptionStyle{ruled}{labelfont=normalfont,labelsep=colon,strut=off} 
\floatstyle{ruled}
\newfloat{listing}{tb}{lst}{}
\floatname{listing}{Listing}
\title{Attending to Routers Aids Indoor Wireless Localization%
\thanks{Accepted at the AAAI 2026 Workshop on Machine Learning for Wireless Communication and Networks (ML4Wireless)}}
\author{
    Ayush Roy\thanks{Equal contributions.},
    Tahsin Fuad Hassan\textsuperscript{\dag},
    Roshan Ayyalasomayajula,
    Vishnu Suresh Lokhande
}
\affiliations {
    Department of Computer Science and Engineering, University at Buffalo-SUNY\\
    aroy25@buffalo.edu, tahsinfu@buffalo.edu, roshana@buffalo.edu, vishnulo@buffalo.edu.
}

\begin{document}

\maketitle

\begin{abstract}
Modern machine learning-based wireless localization using Wi-Fi signals continues to face significant challenges in achieving groundbreaking performance across diverse environments. A major limitation is that most existing algorithms do not appropriately weight the information from different routers during aggregation, resulting in suboptimal convergence and reduced accuracy. Motivated by traditional weighted triangulation methods, this paper introduces the concept of attention to routers, ensuring that each router’s contribution is weighted differently when aggregating information from multiple routers for triangulation. We demonstrate, by incorporating attention layers into a standard machine learning localization architecture, that emphasizing the relevance of each router can substantially improve overall performance.
We have also shown through evaluation over the open-sourced datasets and demonstrate that Attention to Routers outperforms the benchmark architecture by over 30\% in accuracy.
\end{abstract}
\vspace{-1.5em}
{\small
\begin{links}
\textbf{Dataset:}\\ \url{https://github.com/ucsdwcsng/DLoc_pt_code/blob/main/wild.md}{Dataset}\\
\textbf{Code:}\\ \url{https://github.com/AyushRoy2001/Attending-to-Routers}{Code}
\end{links}
\vspace{-1em}}


\section{Introduction}\label{sec:intro}


Indoor localization has accelerated with the increasing deployment of IoT devices and indoor robots~\cite{iot_trend}. Specifically, wireless techniques based on Wi-Fi Channel State Information (CSI)~\cite{kotaru2015spotfi,picoscenes} enable applications in robotics, activity detection, and assistive navigation~\cite{arun2024wais,dloc,zhang2024rloc}, driving a projected $43.2$~bn market by 2030. Accessibility has improved with the development of open-source toolboxes~\cite{picoscenes,arun2024wais} and advances in Wi-Fi standards~\cite{du2024overview}. Over time, localization solutions have moved from RSSI-based~\cite{radar} to CSI-based methods~\cite{chronos,kotaru2015spotfi}, with recent data-driven approaches~\cite{dloc,zhang2024rloc} that address Non-Line of Sight (NLoS) challenges.

Current machine learning architectures~\cite{dloc,zhang2024rloc} often assign equal weight to routers, or expect the network to infer the weights implicitly based on aggregate performance. This practice increases the number of parameters and can reduce localization accuracy \cite{nazarovs2021graph}. In contrast, traditional localization algorithms~\cite{kotaru2015spotfi} improve performance with weighted triangulation, where router weights are hand-tuned. Within machine learning models, these weights can be optimized using attention mechanisms~\cite{vaswani2017attention,lee2019set}.

In this work, we investigate the benefits of integrating attention layers into baseline machine learning models for localization, allowing the model to explicitly learn the importance of each router a concept we refer to as \textit{Attention to Routers}. This explicit weighting improves the network’s performance on the localization objective.

\begin{figure}[t]
    \centering
    \includegraphics[width=0.55\linewidth]{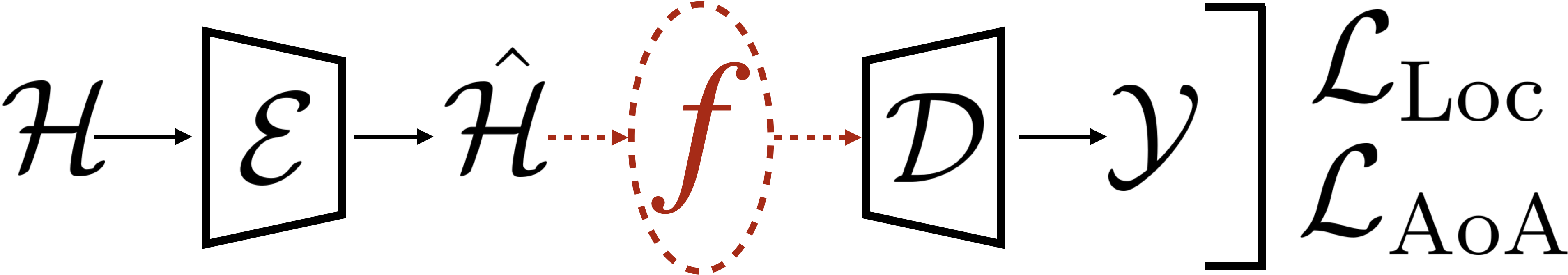}
    \caption{The input heatmaps $\mathcal{H}$ are encoded by $\mathcal{E}$, processed by set-invariant attention $f$, and decoded by $\mathcal{D}$ to predict $y$, supervised with losses $\mathcal{L}_{\mathrm{Loc}}$ and $\mathcal{L}_{\mathrm{AoA}}$.}
    \vspace{-1.5em}
    \label{fig:main_architecture}
\end{figure}

We evaluate Attention to Routers by constructing a baseline model and an enhanced version with attention layers. In our baseline, we follow DLoc~\cite{dloc} and RLoc~\cite{zhang2024rloc}, utilizing Angle-of-Arrival and Time-of-Flight (AoA-ToF) heatmaps from each router to extract accurate AoA values. These are used for triangulation, supported by a triangulation loss. To introduce a straightforward attention mechanism, we apply attention over the encoder’s embeddings in the encoder-decoder model (see Figure~\ref{fig:main_architecture}), as these embeddings summarize the input AoA-ToF heatmaps.

Finally, we compare our algorithm against a baseline machine learning model without attend-and-excite, using one of the most widely used open-source datasets provided by DLoc~\cite{dloc}. Our results, benchmarked against a vanilla machine learning algorithm, show significant improvements:

i) Localization error, trained and tested across the environment, is $44$~cm (median) and $94$~cm ($90^{th}$ percentile), outperforming the baseline by $30\%$.

ii) Attention models reduce baseline errors at harder locations by $45\%$ and at moderately difficult locations by $26\%$, highlighting the advantage of Attention to Routers.

\section{An Encoder-Decoder WiFi Localization}
Let $\mathcal{H} = [H_{1}, H_{2}, \dots, H_{N_{AP}}]$ denote the stack of two-dimensional heatmaps obtained from $N_{AP}$ access points (APs). Each heatmap $H_{i}$ encodes the likelihood distribution in relative polar $(AoA,ToF)$ coordinates of the client device with respect to AP locations  $\mathcal{R} = [r_1,r_2,\cdots,r_{N_{AP}}]$, after converting raw channel state information (CSI) through angle-of-arrival (AoA) and time-of-flight (ToF) processing~\cite{dloc}. The ground-truth target for the network is another image $\mathcal{T}_{\text{location}}$ of identical spatial dimensions, in which the true client position is represented as a Gaussian peak rather than a one-hot pixel. This smooth representation improves gradient flow during training and mitigates vanishing gradient issues. Consequently, the localization task is formulated as an image-translation problem, mapping $\mathcal{H} \rightarrow \mathcal{T}_{\text{location}}$, which facilitates generalization to arbitrary environmental layouts and AP deployments~\cite{dloc}.

\paragraph{Network Design.}
The model consists of a single encoder-decoder architecture \cite{lokhande2022equivariance}, where the encoder $\mathcal{E}\colon \mathcal{H} \rightarrow \hat{\mathcal{H}}$ compresses the multi-AP input stack into a latent feature representation $\hat{\mathcal{H}}$, and the decoder $D\colon \hat{H} \rightarrow Y$ reconstructs a spatial likelihood map of the client’s position. The architecture is inspired by ResNet-based image translation networks~\cite{dloc}. The encoder begins with a $7 \times 7$ convolution followed by a \texttt{Tanh} activation to mimic log-scale feature combination, while subsequent layers employ residual blocks for hierarchical representation learning. The decoder mirrors this structure with transposed convolutions, instance normalization, and \texttt{ReLU} activations to recover spatial resolution and generate the output heatmap $\mathcal{Y}$. This design enables the network to implicitly account for environment geometry, multipath reflections, and random ToF offsets across APs.

\paragraph{Loss Function.}
Given predicted output $\mathcal{Y} = \mathcal{D}(\mathcal{E}(\mathcal{H}))$ and target $\mathcal{T}_{\text{location}}$, the training objective combines pixel-wise reconstruction and sparsity regularization. The primary term is an L2 loss enforcing similarity to the 2D-location target:
\begin{equation}
    \mathcal{L}_{\text{Loc}} = \| Tri(\mathcal{D}(\mathcal{E}(\mathcal{H})), \mathcal{R}) - \mathcal{T}_{\text{location}} \|_2^2,
\end{equation}
where $Tri(\cdot, \cdot)$, is the standard triangulation algorithm that uses the AoA values predicted by the network along with the router locations $\mathcal{R}$. Because the desired output is the final location predicted from the Angle of Arrival (AoA), we also enforce an L1 loss over the AoA values predicted by the decoder for each AP's embeddings:
\begin{equation}
    \mathcal{L}_{\text{AoA}} = \lambda \, \| \mathcal{D}(\mathcal{E}(\mathcal{H}) - \mathcal{T}_{\text{AoA}} \|_1.
\end{equation}
The total loss is thus
\begin{align}
    \mathcal{L} = \mathcal{L}_{\text{Loc}} + \mathcal{L}_{\text{AoA}}
    & = \| Tri(\mathcal{D}(\mathcal{E}(\mathcal{H})), \mathcal{R}) - \mathcal{T}_{\text{location}} \|_2^2 \\ \nonumber
    & \quad+ \| \mathcal{D}(\mathcal{E}(\mathcal{H})) - \mathcal{T}_{\text{AoA}} \|_1,
\end{align}
where $\lambda$ is a tunable weight controlling the sparsity strength. This objective jointly optimizes the encoder and decoder to produce precise and accurate client location~\cite{dloc}.

\section{Channel-wise Attention for Router Importance Weighting}

Recall that our architecture so far includes an encoder $\mathcal{E}: \mathcal{H} \rightarrow \hat{\mathcal{H}}$ and a decoder $\mathcal{D}: \hat{\mathcal{H}} \rightarrow \mathcal{H}$. Given that the input tensor $\mathcal{H}$ comprises multiple channels corresponding to routers (access points, APs), it is natural to assume that not all routers contribute equally to localization accuracy. 
Routers subject to multipath interference or weaker signals tend to introduce noise, whereas others provide reliable spatial cues. To model this heterogeneity, we introduce a lightweight \emph{channel-wise attention mechanism} \cite{vaswani2017attention,lee2019set} that adaptively emphasizes informative routers while attenuating unreliable ones. This module is inserted between the encoder and decoder and modifies the latent representation $\hat{\mathcal{H}}$ before decoding.

\subsection{Set-Invariant Functional Attention}
Let $\hat{\mathcal{H}} = [\hat{h}_1, \hat{h}_2, \dots, \hat{h}_R]$ denote the encoded embeddings from $R$ routers, where $\hat{h}_r \in \mathbf{R}^d$ represents  $d$-dimensional feature embedding for router $r$. We seek a mapping 
\begin{align}
f: \{\hat{h}_1, \hat{h}_2, \dots, \hat{h}_R\} \rightarrow \{\alpha_1, \alpha_2, \dots, \alpha_R\},
\end{align}
where $\alpha_r \in [0,1]$ and $\sum_{r=1}^R \alpha_r = 1$, such that routers with higher localization relevance receive larger $\alpha_r$. The function $f$ is drawn from the space of \emph{set-invariant functionals} $\mathcal{F}$ \footnote{See Bloem-Reddy and Teh, \emph{Probabilistic Symmetries and Invariant Neural Networks}, Sec.~2.1 and Example~1 (``Deep Sets'') for formal definitions of functional symmetry and set-invariant functionals. Example~1 characterizes set-invariant functions and gives the canonical representation $f(X_n)=\rho\!\big(\sum_i \phi(X_i)\big)$. \citep[Sec.~2.1]{bloem2020probabilistic}}, meaning that $f$ is invariant to permutations of its inputs formally, $f(\{\hat{h}_{\pi(1)}, \dots, \hat{h}_{\pi(R)}\}) = f(\{\hat{h}_1, \dots, \hat{h}_R\})$ for any permutation $\pi$. This ensures that router attention depends on their representations rather than their ordering.

\subsection{Computation of Attention Weights}
To compute $\alpha_r$, we first summarize each router embedding via average pooling:
\begin{align}
s_r = \frac{1}{d}\sum_{j=1}^{d} \hat{h}_{rj}, \quad \forall r \in \{1, \dots, R\}
\end{align}
resulting in a summary vector $\mathbf{s} = [s_1, s_2, \dots, s_R]^\top$. A lightweight multilayer perceptron (MLP) $g(\cdot)$ then projects each $s_r$ into a scalar attention score $u_r$:
\begin{align}
u_r = g(s_r) = W_2 \, \sigma(W_1 s_r + b_1) + b_2,
\end{align}
where $\sigma(\cdot)$ denotes a ReLU activation, and $W_1, W_2, b_1, b_2$ are trainable parameters. These unnormalized scores are converted into probabilistic attention weights via a Softmax operation:
\begin{align}
\alpha_r = \frac{\exp(u_r)}{\sum_{k=1}^R \exp(u_k)}
\end{align}
The Softmax acts as a \emph{self-gating} mechanism \cite{hu2018squeeze}, introducing non-linearity and ensuring that attention weights are differentiable and normalized. Optionally, a global context vector $\bar{h} = \frac{1}{R}\sum_{r=1}^R \hat{h}_r$ can be concatenated to each $\hat{h}_r$ before scoring to enable relative comparison between routers.

\subsection{Feature Recalibration and Interpretability}
Finally, the attention weights $\alpha_r$ are used to recalibrate the latent features:
\begin{align}
\tilde{h}_r = \alpha_r \cdot \hat{h}_r, \quad \forall r \in \{1, \dots, R\}
\end{align}
resulting in the attended embedding $\tilde{\mathcal{H}} = [\tilde{h}_1, \tilde{h}_2, \dots, \tilde{h}_R]$. This attended feature map is passed to decoder $\mathcal{D}$ for reconstruction or localization prediction. The mechanism effectively amplifies embeddings from routers that exhibit stable and informative signal patterns while suppressing those dominated by noise or multi-path distortion. Moreover, learned attention weights $\{\alpha_r\}$ offer interpretable measure of each router’s contribution to localization, providing valuable insights into spatial relevance of network layout.

\section{Experiments}

\subsection{Localization Error in Easy and Hard Cases}

\begin{figure*}
\centering
\begin{minipage}{0.33\textwidth}
     \centering
    \includegraphics[width=\linewidth]{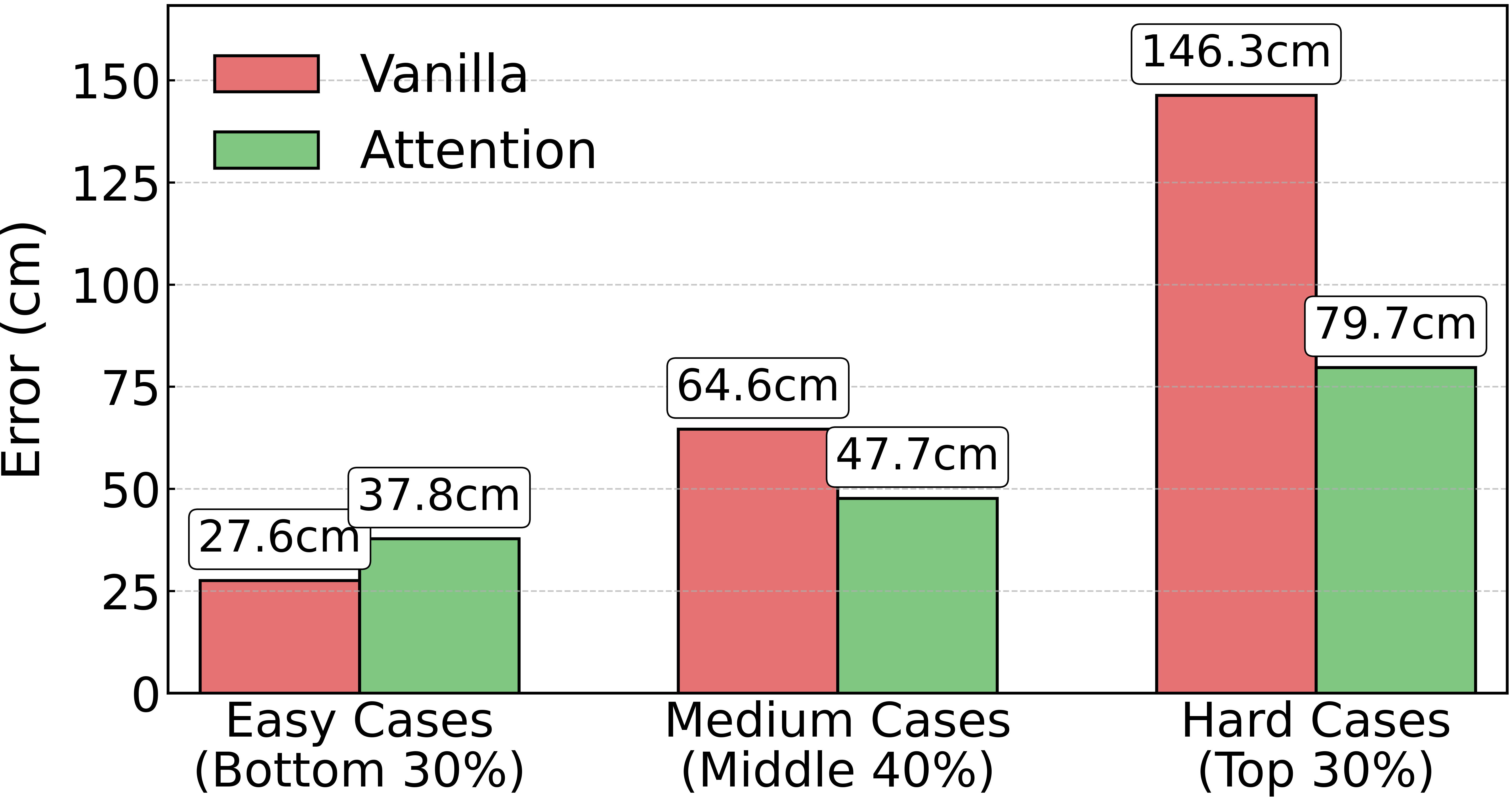}
\end{minipage}
\begin{minipage}{0.33\textwidth}
     \centering
    \includegraphics[width=\linewidth]{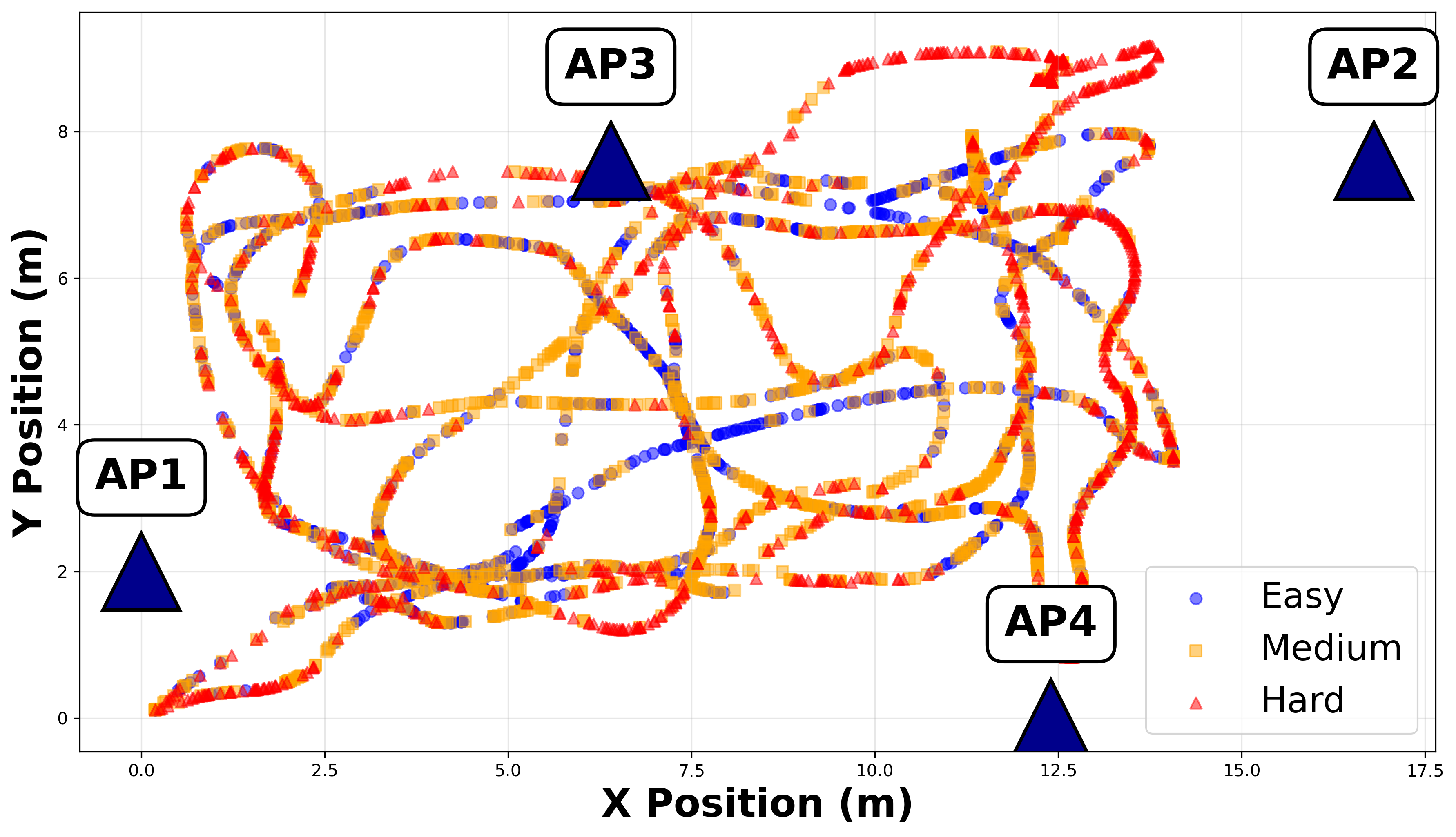}
\end{minipage}
\begin{minipage}{0.3\textwidth}
     \centering
    \includegraphics[width=\linewidth]{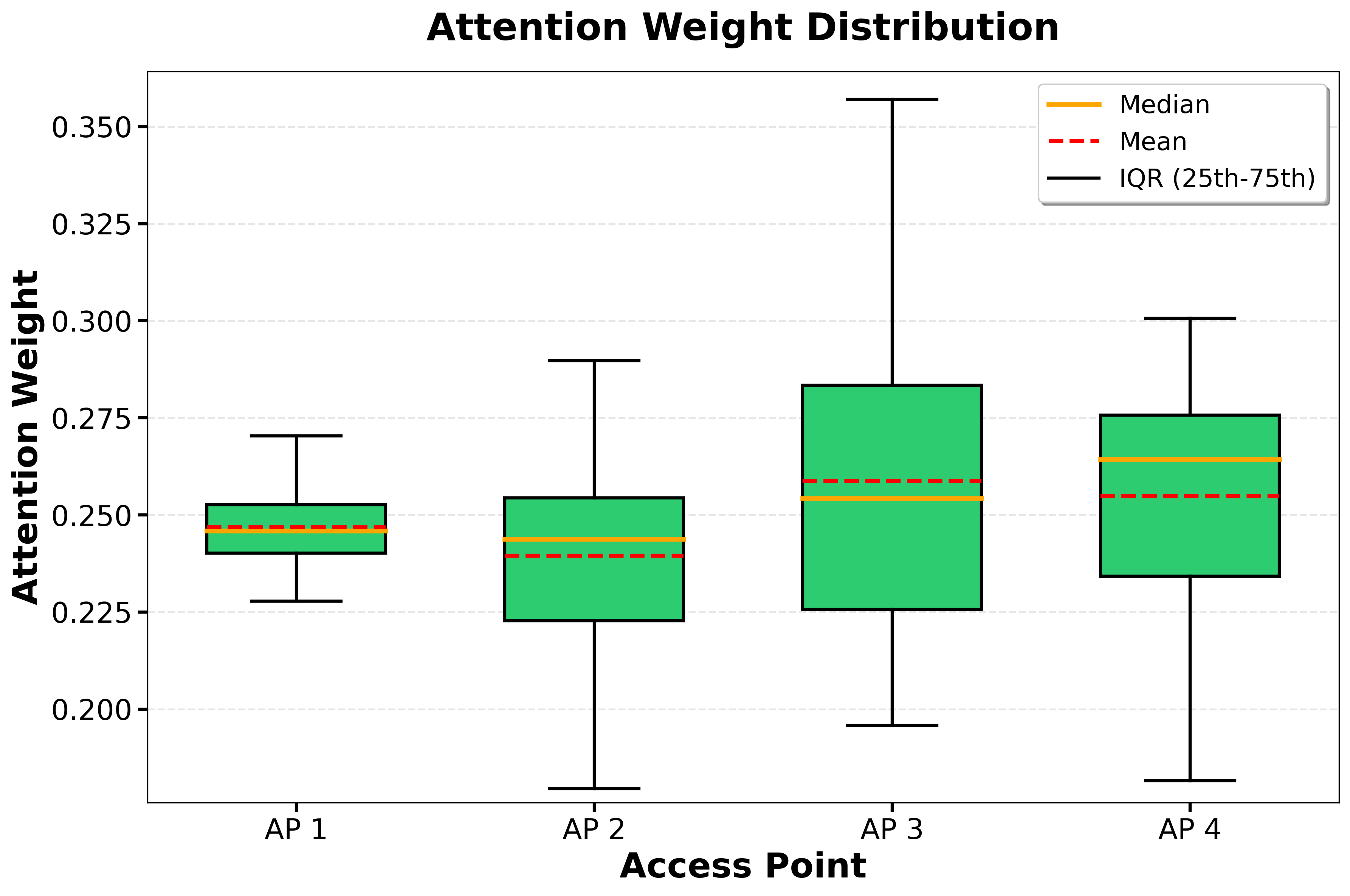}
\end{minipage}
\caption{\footnotesize(a) \textbf{Performance comparison between Vanilla and Attention models across case difficulty levels:}  The mean localization error (in cm) is reported for three difficulty tiers: Easy (bottom $30\%$), Medium (middle $40\%$), and Hard (top $30\%$) cases.
(b) \textbf{Spatial distribution of easy vs medium vs hard cases overlaid with AP locations:} Easy (blue), Medium (orange) and Hard (red) samples are plotted with counts in the legend; APs are labeled and emphasized to show which access points are surrounded by high-error samples. (c) \textbf{Router based Attention weight distribution:} Shows the attention weights' distribution for each AP. We can clearly see tighter distributions for APs that provide equal attention to all samples like AP1, and at AP4 which has more skewed distribution demonstrating skewed attention across samples.
    }
    \vspace{-1.5em}
    \label{fig:results}

\end{figure*}

Figure \ref{fig:results}(b) illustrates the spatial distribution of easy, medium, and hard samples relative to the Access Points (APs). Here, easy samples correspond to locations with low localization error, medium samples exhibit moderate error, and hard samples represent high-error or ambiguous cases. The plot clearly reveals where these categories cluster in the environment, showing that hard cases often concentrate around specific APs (highlighted on the map), whereas easy cases tend to appear in regions with denser AP coverage. This pattern suggests that the attention mechanism learns to allocate greater representational capacity to spatially challenging or under-determined regions, thereby mitigating the impact of unreliable AP geometry. Table \ref{tab:localization_comparison} provides a quantitative comparison of localization error between the baseline and our attention-aided model across multiple statistical measures. Consistent improvements are observed at all percentiles, with the proposed approach achieving a $28.7\%$ reduction in median error and up to $39.4\%$ improvement at the $99^\text{th}$ percentile. Figure \ref{fig:results}(a) further breaks down the mean localization error by difficulty category: while the attention-based model shows a modest $36.9\%$ increase in error for easy cases, reflecting a deliberate redistribution of representational focus, it achieves substantial gains of $26.2\%$ and $45.5\%$ for medium and hard cases, respectively. Collectively, these results indicate that the attention mechanism enhances robustness by emphasizing complex and ambiguous samples, reducing high-error outliers, and maintaining overall performance stability across varying environmental conditions.

\begin{table}[h]
\centering
\caption{Comparison of localization error between baseline and attention-based method across different statistical measures. Results are based on 3,966 total samples, showing consistent improvement across all percentiles with the attention-based approach. The method achieves a $28.7\%$ reduction in median error ($18.16$ cm improvement) and up to $39.4\%$ improvement at the $99^\text{th}$ percentile.}
\label{tab:localization_comparison}
\begin{tabular}{lccc}
\hline
\textbf{Metric} & \textbf{Base (cm)} & \textbf{Ours (cm)} & \textbf{$\Delta$} \\
\hline
Median & 63.17 & 45.01 & +28.7\% \\
Mean & 77.90 & 54.01 & +30.7\% \\
90th Percentile & 140.63 & 92.88 & +34.0\% \\
95th Percentile & 172.00 & 114.32 & +33.5\% \\
99th Percentile & 302.32 & 183.20 & +39.4\% \\
\hline
\end{tabular}
\vspace{-1.0em}
\end{table}


\subsection{Weights of the Attention mechanism}


Figure 2(c) visualizes the distribution of attention weights assigned to each Access Point (AP). Each box represents the interquartile range (IQR) of attention weights across all test samples, with the mean and median indicated by dashed red and solid yellow lines, respectively. The model distributes focus non-uniformly across APs, highlighting the learned spatial selectivity of the attention mechanism. In particular, AP~$3$ receives the highest median and mean attention, indicating its stronger relevance for accurate localization in the given environment. Conversely, AP~$1$ and AP~$2$ exhibit tighter distributions around lower weights, suggesting consistent but lower contribution. The overall entropy of the mean attention vector corresponds to approximately $92\%$ uniformity, confirming that while the model leverages all APs, it adaptively prioritizes the most informative ones.

\subsection{Localization Error CDF analysis}


Table~\ref{tab:localization_comparison} presents the various percentiles of errors from the cumulative distribution function (CDF) of localization error for the baseline model and the proposed attention model. Across full range of error magnitudes, the attention-enhanced variant consistently achieves higher cumulative probabilities, indicating that a larger fraction of samples attain lower localization error. Most notable improvement appears in medium-to-high error regime, where the green curve lies distinctly above the baseline, reflecting superior robustness under challenging conditions. Quantitatively, median error ($50^\text{th}$ percentile) decreases from approximately $59$ cm to $45$ cm, while the $90^\text{th}$ percentile drops from $120$ cm to about $100$ cm, corresponding to a $28.7\%$ and $39.4\%$ reduction, respectively. These shifts confirm the attention module does not merely improve average accuracy but effectively suppresses extreme outlier errors. Shaded region between the curves visualizes this consistent gain across all percentiles, demonstrating that the attention module improves both reliability and stability in localization.

\section{Discussion}\label{sec:discussion}

In this work, we introduce attention mechanisms into machine learning-based Wi-Fi indoor localization models, allowing the network to learn and emphasize the contribution of each router. This approach, inspired by weighted triangulation in traditional localization, yields substantial improvements in convergence and accuracy, particularly under challenging conditions such as Non-Line of Sight and multipath scenarios. Attention weights provide interpretability by prioritizing routers with more reliable signals and mitigating errors from noisier sources, resulting in up to 39.4\% error reduction at the 99th percentile and significant improvements for hard and moderately difficult locations. By enabling adaptive weighting within encoder-decoder frameworks, our method enhances robustness, mitigates high-error outliers, and supports more generalizable and resilient localization performance in complex indoor environments.


\bibliography{aaai2026}

\end{document}